\documentclass[conference]{IEEEtran}
 \IEEEoverridecommandlockouts

\usepackage{times}
\usepackage{epsfig}
\usepackage{graphicx}
\usepackage{amsmath}
\usepackage{amssymb}

\usepackage{url}
\usepackage{hyperref}
\usepackage{amsfonts}
\usepackage{bbm}
\usepackage{booktabs}
\usepackage{algorithm}
\usepackage{algorithmic}
\usepackage[export]{adjustbox}
\usepackage{subfigure}
\usepackage{bbding}

\begin{document}
\title{ Wheat Crop Yield Prediction Using Deep LSTM Model}

\author{\IEEEauthorblockN{Sagarika Sharma\IEEEauthorrefmark{1}\thanks{\IEEEauthorrefmark{1}Authors with equal contribution.}\IEEEauthorrefmark{2}, Sujit Rai\IEEEauthorrefmark{1} \IEEEauthorrefmark{3}, Narayanan C. Krishnan \IEEEauthorrefmark{4}}
\IEEEauthorblockA{\textit{Indian Institute of Technology Ropar, India} \\
\IEEEauthorrefmark{2}2017csb1012@iitrpr.ac.in, \IEEEauthorrefmark{3}2017csm1006@iitrpr.ac.in,
\IEEEauthorrefmark{4}ckn@iitrpr.ac.in
}
}

\maketitle

\begin{abstract}
An in-season early crop yield forecast before harvest can benefit the farmers to improve the production and enable various agencies to devise plans accordingly. We introduce a reliable and inexpensive method to predict crop yields from publicly available satellite imagery. The proposed method works directly on raw satellite imagery without the need to extract any hand-crafted features or perform dimensionality reduction on the images. The approach implicitly models the relevance of the different steps in the growing season, and the various bands in the satellite imagery.  We evaluate the proposed approach on tehsil (block) level wheat predictions across several states in India and demonstrate that it outperforms over existing methods by over 50\%. We also show that incorporating additional contextual information such as the location of farmlands, water bodies, and urban areas helps in improving the yield estimates.

 
\end{abstract}
\section{Introduction}
India is predominantly an agrarian economy, with nearly 70\% of its rural households depending primarily on agriculture for their livelihood. Approximately 82\% of farmers are small and marginal landowners (FAO, India at a glance~\cite{FAO}), which is estimated to grow to 91\% by 2030.  The marginal landholdings combined with traditional modes of farming, and external factors such as irregular rainfall, depletion of groundwater, etc., result in the yield of crops in India still being below the world average (Economic Survey 2015-16~\cite{unionbudget}). Existing approaches for estimating yield rely on manual surveys during the growing season. The enormous costs and the manual effort required to conduct such studies makes it a cumbersome method to predict crop yields. At the same time, lack of reliable and up to date information on crop yield affects supply-demand stocks and export options. Thus, an in-season crop yield forecast can benefit the farmers to improve production and enable the government agencies to devise appropriate plans.

Remote sensing data is becoming an increasingly popular source of data for developing models for various applications such as {poverty estimation}~\cite{xie}, {income prediction}~\cite{tushar}, {yield estimation}~\cite{aaai}, etc. The easy and inexpensive access, high-resolution imagery combined with increasing sophisticated modeling techniques, makes it a viable solution for many of these problems. In particular, multi-spectral satellite imagery have information across a wide spectrum of wavelengths that abundantly encode information related to land use such as vegetation, water bodies, urban areas, etc. 

In this paper, we propose an approach for crop yield estimation from satellite imagery using deep learning techniques that have found success in traditional computer vision tasks. Unlike prior methods that involve extracting hand-crafted features or rudimentary features such as histograms, our approach directly works on the satellite images. It allows the model to learn the representations that are useful for the yield prediction task. Often yield estimates in a geographical area depend on other factors such as nearby water bodies, urbanization, etc. While prior approaches to yield prediction do not take into account these factors, we incorporate these aspects by explicitly weaving into our model the land use classification data. We model the temporal features in the data through a deep LSTM model. This allows our model to automatically identify the relevance of the different growing steps and the satellite image bands towards the yield prediction task. 

We evaluate and validate our approach on the task of tehsil (block) level wheat prediction for seven states in India. We use the MODIS surface reflectance multi-spectral satellite images, along with the land use classification maps, to train the proposed deep learning models. The experimental results show that our model can outperform traditional remote-sensing based methods by 70\% and recently introduced deep learning models by 54\%. To the best of our knowledge, this is the first method that yields promising results for crop yield prediction in the Indian context.

\section{Related Work}
In recent years, remote sensing data has been widely used in various sustainability applications such as land use classification ~\cite{albert}, infrastructure quality prediction ~\cite{kdd}, poverty estimation ~\cite{xie}, population estimation ~\cite{population} and income level predictions ~\cite{tushar}.

Crop yield estimation using remote sensing data has also been explored over the past few years. Prasad et al. ~\cite{anoop} employ a piece-wise linear regression and a non-linear Quasi-Newton multi-variate regression model to predict soybean yield in the state of Iowa using normalized difference vegetation index (NDVI), soil moisture, surface temperature, and rainfall data. Kuwata and Shibasaki ~\cite{kuwata} estimate county-level crop yield for the state of Illinois using MOD09A1 derived EVI (Enhanced Vegetation Index), climate and other environmental data by employing deep neural network and SVM. Johnson et al. ~\cite{jonson} learn models to predict corn and soybean yields from NDVI and daytime land surface temperature data (derived from the Aqua MODIS sensor product MYD11A2) using a regression tree. Mallick et al. ~\cite{rice} use the Vegetation Condition Index (VCI) that is derived from NDVI and Normalised Difference Wetness Index (NDWI) for rice yield prediction in India, while Dubey et al.  ~\cite{sugarcane-fasal} use VCI to model sugarcane yield variability in 52 Indian districts. The Indian national-level program, called FASAL (Forecasting Agriculture using Space, Agro-meteorology, and Land-based observations), has been operational since 2006. FASAL aims at providing pre-harvest crop production forecasts at National/State/District level ~\cite{ray14a}. However, information about the forecasts is scarcely available in the public domain.

All these prior approaches learn to model yield using some form of vegetation index that is derived from multi-spectral satellite imagery rather than directly employ the satellite imagery.
These approaches mostly utilize 2 or 3 bands that are traditionally used in the generation of these indices. In contrast, we let our model to automatically learn the utility of the different bands during the crop growing season. The model also learns to implicitly estimate the importance of the multi-spectral satellite images belonging to different phases in the crop growing season. 

Our study is inspired by the work of You et al.~\cite{aaai} on corn yield prediction using remote sensing data and deep learning models. You et al. extract histograms of crop pixel intensities estimated using crop masks on multi-spectral satellite images. The series of histograms obtained during the growing season of the corn crop is modeled using LSTM and Gaussian processes to predict the crop yield.  In contrast, our approach works directly on the raw multi-spectral satellite images to learn the representations that are crucial for crop yield prediction. We also incorporate additional information on nearby water bodies and urban built-up to train deep neural network models that yield better results. 
\begin{figure}[!t]
\centering
\includegraphics[width=0.45\textwidth]{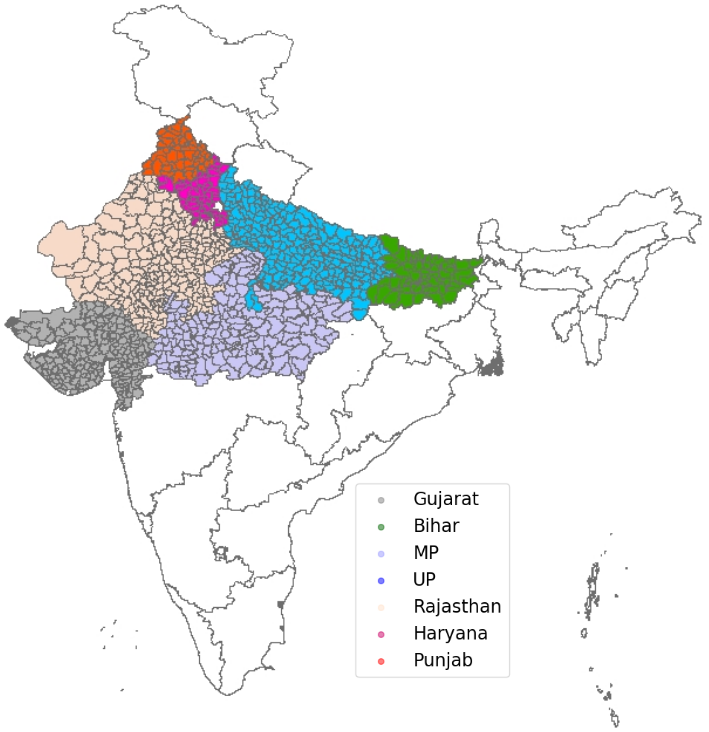}
\caption{Study Area with all the 948 tehsils belonging to 7 states of India}
\label{fig:studyarea}
\end{figure}

\section{Problem Setting and Data Description}
The primary objective of the work is to build a crop yield prediction model for the wheat crop using multi-spectral satellite imagery. A series of satellite images during the growing season of wheat before the harvest is given as input to the model. Wheat is typically grown and harvested in the Rabi season (October-April) in India. Hence we focus on this growing period for learning the model. 

We have collected the statistics of crop yield from the open government data platform~\cite{OGD}. In India, the lowest administrative unit for which the statistics of crop yield are available is at the district level. However, the average size of a satellite image required to cover a district would be too large for training the model ($>1024\times1024)$. We, therefore, split the yield of a district across smaller administrative units called tehsils, taking into account the agricultural area in each tehsil. A tehsil (also known as a Mandal, or taluk) is an administrative division in India that comprises of multiple villages in the rural areas and various blocks in urban areas. The maximum size of a satellite image required to cover a tehsil is 300$\times$300. Predicting the crop yield at the tehsil level will also help the agencies to device customized plans for improved utilization of resources. 

In this study, we focus on the seven major wheat growing states that together account for more than 90\%of the total wheat production in India. The crop yield data for these states at the district level are only available from 2001-2011. There are a total of 948 tehsils in our study, with a tehsil having an average geographical spread of over 35,000 hectares. The state-wise distribution of these tehsils and the average wheat crop yield for the year 2011 is provided in Table \ref{tab:statistics}. The geographical spread of the study area is illustrated in Figure \ref{fig:studyarea}.
\begin{table}
\centering
\small
\caption{Statistics of tehsils and wheat crop yield for the year 2011 in the dataset.}
\begin{tabular} {|l|c|p{1.5cm}|p{1.5cm}|}
 \hline
 \hline
 State & No. of tehsils & Average area (in hectares) & Average yield (kgs/hectare)\\
 \hline
 \hline
 Gujarat & 215 & 7040.0 & 505.3 \\
 \hline
 Bihar & 53 & 39123.9 & 2001.8\\
 \hline
 Haryana & 46 & 51392.9 & 2354.5\\
 \hline
 Madhya Pradesh & 167 & 28283.4 & 698.5\\
 \hline
 Uttar Pradesh & 209 & 44361.6 & 1133.2\\
 \hline
 Rajasthan & 211 & 14588.6 & 5768.6\\
 \hline
 Punjab & 47 & 67125.0 & 2404.7\\
 \hline
 \hline
\end{tabular}
\label{tab:statistics}
\end{table}

The proposed work uses publicly available satellite data from the following MODIS sensors onboard NASA's Terra and Aqua satellites~\cite{lpdaac}:
\begin{itemize}
    \item MOD09A1-  This is also referred to as the MODIS  Surface Reflectance 8-Day L3 Global product. It provides an estimate of the surface spectral reflectance as it would be measured at ground level in the absence of atmospheric scattering or absorption with a spatial resolution of 500m. 
    \item MYD11A2- It is an eight-day composite thermal product from the Aqua MODIS sensor.
    \item MODIS Land Cover -The primary land cover scheme incorporated by the MODIS Terra+Aqua Combined Land Cover product identifies 17 classes defined by the IGBP(International Geosphere-
Biosphere Programme), including 11 natural vegetation classes, three human-altered classes, and three non-vegetated classes with a spatial resolution of 500m. A pixel is assigned to a class if 60\% or more of the area covered by the pixel belongs to the class. In our study, we only consider pixels that have been classified as agriculture, water bodies, and urban built-up.
\end{itemize}
\begin{figure}[!t]
\centering
\includegraphics[width=0.45\textwidth]{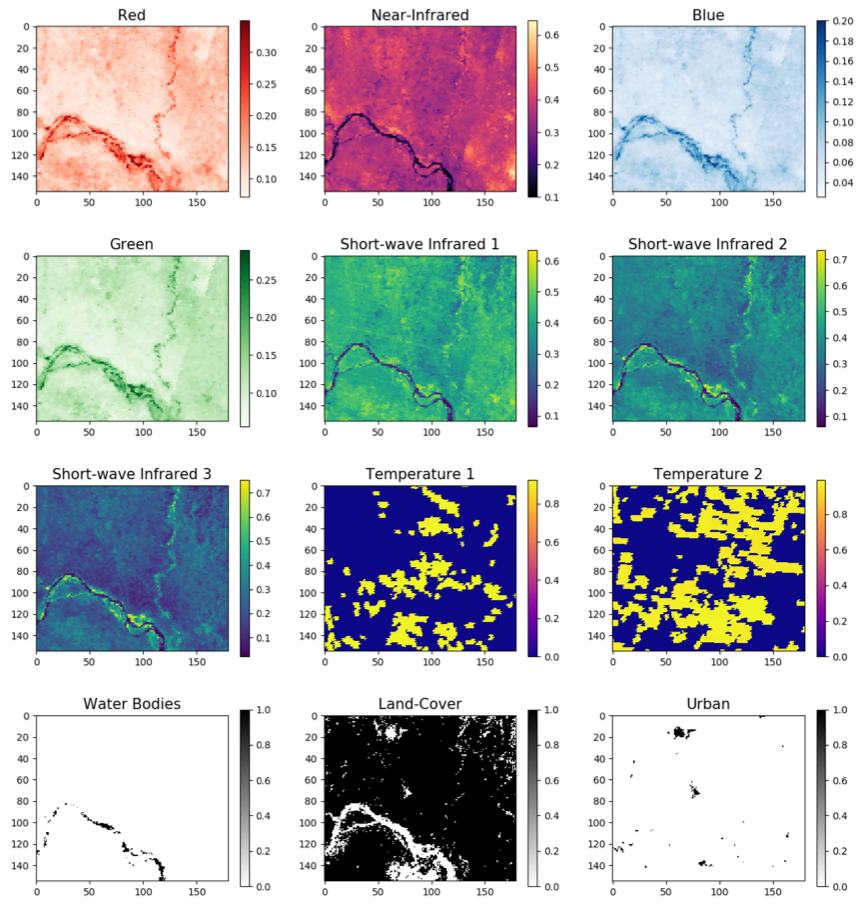}
\caption{[Best Viewed in Color] Visualization of the different satellite image bands for a tehsil.}
\label{fig:bands}
\end{figure}
\begin{figure*}
\centering
\includegraphics[width=0.8\textwidth]{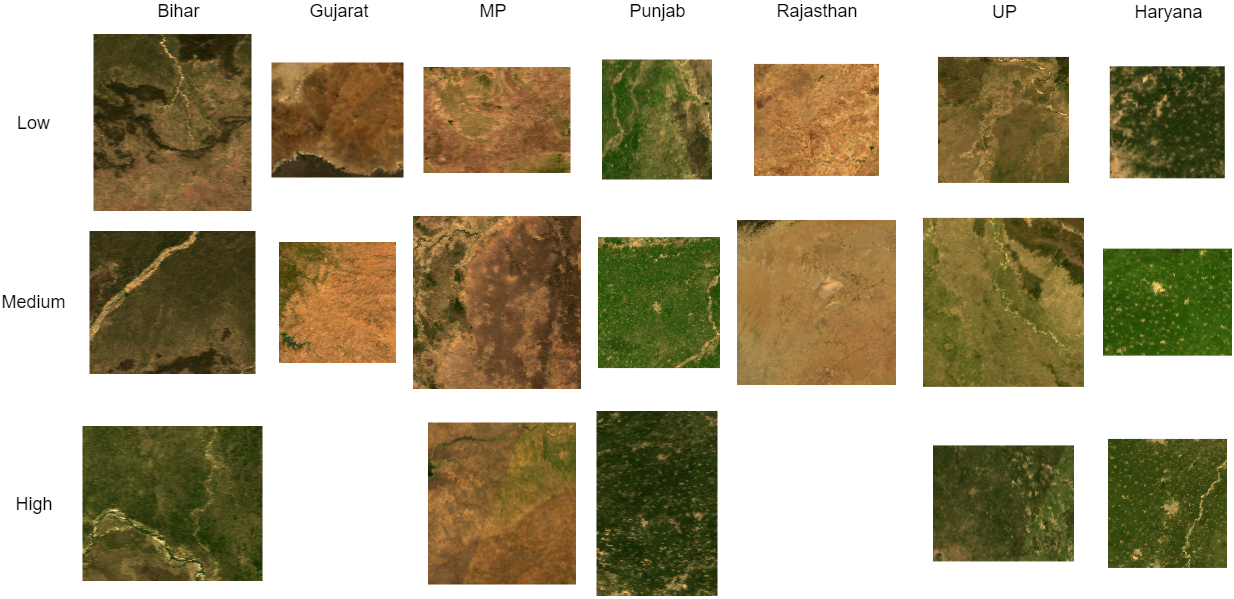}
\caption{[Best Viewed in Color] Visual images for different yield levels for the seven states}
\label{fig:variationinstates}
\end{figure*}
Each multi-spectral satellite image, $S_t$, consists of 7 bands of MODIS land surface reflectance image MOD09A1, two bands of MODIS land surface temperature, and three binary bands derived from MODIS land cover image corresponding to water bodies, agricultural land and urban built-up. These bands are illustrated in Figure \ref{fig:bands} for a tehsil from the state of Bihar. Prior approaches use vegetation indices derived mostly from the bands 1 and 2.

Figure \ref{fig:variationinstates} illustrates representative visual images for low, medium, and high yielding tehsils for each of the seven states. A significant variation in these images is observed across all the states. The variance is to such an extent that a couple of states do not have a high crop yielding tehsil. Further, we observe a lot of variation in the vegetation land space across tehsils that are supposed to result in a similar yield. For example, tehsils with medium yield in Punjab and Haryana appear to be a lot greener than states such as Rajasthan and Gujarat. This level of heterogeneity in the data made us decide to model the yield in each state independently. We also show through our experiments the difficulty in predicting the yield of state using a model that has been trained on the data from a different state.

\section{Methodology}

\subsection{Preliminaries}
We first give a brief overview of the deep neural network models that are the building blocks of our crop yield estimator before describing the final model architecture.
\subsubsection{Deep Convolutional Neural Networks} 
Deep Convolutional Neural Networks (CNN)~\cite{cnn} can be viewed as a large composition of complex nonlinear functions that learn hierarchical representations of the data. A CNN typically consists of two types of layers: fully connected and convolutional layers. A fully connected layer consists of multiple nodes. Each node takes a vector, $\textbf{x}\in \mathcal{R}^D$, as input and outputs a scalar that is a nonlinear transformation of the weighted sum of the inputs in the following manner. 
\begin{equation}
    z = f\left(b + \textbf{w}^T \textbf{x}\right)
\end{equation}
where $\textbf{w}$ are the weights, $b$ is the scalar bias term, and $f(.)$ is the non-linear transformation (usually a rectified linear unit (ReLU) or tanh). 
A convolutional layer typically consists of three main operations: convolution, nonlinear activation, and pooling. The convolution operation is performed using a filter with shared parameters that results in significant reduction in the number of parameters. The filter $\textbf{W} \in \mathcal{R}^{k\times k\times D}$ is convolved with an input tensor $\textbf{X}\in \mathcal{R}^{M\times N \times D}$. These filters are trainable and often learn various local patterns present in the input tensor. The convolution operation is followed by the nonlinear function. ReLU is the popular nonlinear function when working with images. The resulting output can be represented as 
\begin{equation}
    \textbf{Z} = f\left(b + \textbf{W} * \textbf{X}\right)
\end{equation}
where $f$ represents the nonlinear function, and $*$ represents the convolution operation. This is often succeeded by the pooling operation. Pooling can be viewed as a sampling process that summarizes the information present in the input. The most common pooling operation is Max Pooling that outputs the maximum of all inputs within a window of size $k\times k$. The output of the convolutional layer is referred to as a feature map.

Deep CNN has a large number of stacked up convolutional and fully connected layers with the output of one layer acting as the input to the next layer. A large number of layers help CNN learn global patterns present in the input. The weights at each layer are learned using the backpropagation algorithm that follows a standard gradient descent approach to minimizing the overall loss.
\begin{figure*}
\centering
\includegraphics[width=0.8\textwidth]{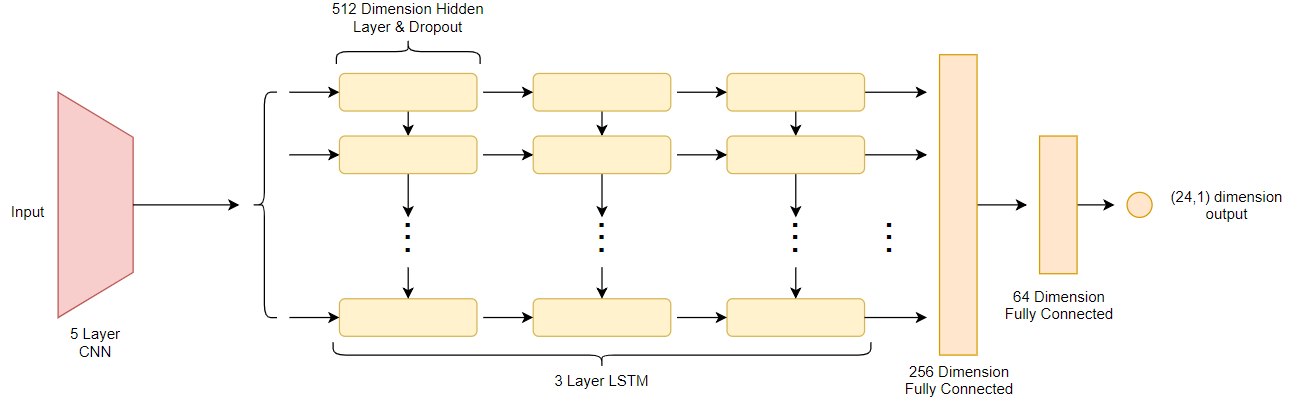}
\caption{The proposed CNN-LSTM architecture for predicting crop yield from a sequence of multi-spectral satellite imagery}
\label{fig:architecture}
\end{figure*}
\subsubsection{Recurrent Neural Networks}
Recurrent neural networks (RNN)\cite{mikolov2010recurrent} are a special type of neural networks for learning sequential data. RNN can remember an encoded representation of its past, thus making it suitable for modeling sequential data. Given a sequential data $\textbf{x}_1, \textbf{x}_2,\ldots, \textbf{x}_T$ for $T$ time steps, the output $\textbf{y}_t$ at time step $t$, is a function of the input at time step $\textbf{x}_t$ and the hidden state $\textbf{z}_{t-1}$ at time step $t-1$, can be defined as follows 
\begin{equation}
    \textbf{z}_t = f(\textbf{w}^T_t \textbf{x}_t+ \textbf{u}^T\textbf{z}_{t-1})
\end{equation}
\begin{equation}
    \textbf{y}_t = g(\textbf{v}^T\textbf{z}_t)
\end{equation}
where, $\textbf{w}$, $\textbf{u}$ and $\textbf{v}$ are the weights applied on $\textbf{x}_t$, $\textbf{z}_{t-1}$ and $\textbf{z}_t$ respectively and $f$ and $g$ are the non-linear activation functions. As, output is dependent on the hidden states of the previous time steps, the back propagation through time algorithm for updating the weights can result in the problem of vanishing or exploding gradients \cite{bengio1994learning}. 

\textbf{LSTM}~\cite{lstm}, a special kind of RNN, were introduced to overcome this issue by integrating a gradient superhighway in the form of a cell state $\textbf{c}$, in addition to the hidden state $\textbf{h}$. The LSTM model has gates for providing the ability to add and remove information to the cell state. The forget gate decides the information to be deleted from the cell state and can be defined as follows
\begin{equation}
    \textbf{f}_t = \sigma (\textbf{w}_f^T [\textbf{h}_{t-1}, \textbf{x}_t] + b_f)
\end{equation}
The input gate that determines the information that should be added to the cell state is defined as 
\begin{equation}
    \textbf{i}_t = \sigma (\textbf{w}_i^T [\textbf{h}_{t-1}, \textbf{x}_t] + b_i)
\end{equation}
The cell state $\textbf{c}_t$ is obtained by using both $f_t$ and $i_t$ in the following manner
\begin{eqnarray}
    \tilde{\textbf{c}}_t & = & tanh(\textbf{w}_c^T [\textbf{h}_{t-1}, \textbf{x}_t] + b_c)\\
     \textbf{c}_t & = & \textbf{f}_t^T \textbf{c}_{t-1} + \textbf{i}_t^T \tilde{\textbf{c}}_t
\end{eqnarray}
Similarly, the hidden state $\textbf{h}_t$ and output state $\textbf{o}_t$ of the LSTM are defined as
\begin{eqnarray}
    \textbf{o}_t & = & \sigma (\textbf{w}_o^T [\textbf{h}_{t-1}, \textbf{x}_t] + b_o)\\
    \textbf{h}_t & = & \textbf{o}_t^T tanh(\textbf{c}_t)
\end{eqnarray}
LSTM is more effective in modeling longer sequences than a simple RNN due to a more effective gradient flow during backpropagation.

\subsection{Crop Yield Prediction Model Architecture}
We directly input the multi-spectral satellite imagery to our deep neural network model. The motivation for using the raw imagery is to be able to extract features relating to the spatial location of crop pixels and the properties of neighboring regions such as water bodies, urban landscapes, etc. We hypothesize that these parameters influence the crop yield.

The proposed deep network has three modules. The first module is a CNN that learns to extract relevant features from the images. The second module is an LSTM that determines the temporal relationship during the crop growing season. The third module is a fully connected network that finally predicts the crop yield. 

The proposed CNN-LSTM architecture is illustrated in Figure \ref{fig:architecture}. The input to the network is a sequence $S_1, S_2,...,S_{24}$, where $S_t$ is a multi-spectral image of size $300 \times 300 \times b$ at time $t$, where $b$ refers to the number of bands. In the proposed model, we use 12 bands. The entire sequence is used during training and validation. During testing, we vary the sequence length between 1 and 23.

The image $S_t$ at every time step $t$ is first passed to the CNN feature extractor to extract the features $f_t^s$ present in the image. The CNN feature extractor consists of 5 convolutional layers, each having 16 filters of size $[3\times 3]$ with a stride size of $[2\times 2]$ and Leaky-ReLU as the activation function. The choice of the number of convolutional layers and filters in each layer was constrained by the computational resources. There is no pooling operation due to the use of strided convolutions. The output of the convolutional feature extractor is flattened into a 1024 dimensional vector. The features extracted for each of the $T$ time steps are stacked and passed on to the LSTM model.

The LSTM model is used to encode the temporal properties across the growing season. The model consists of 3 LSTM layers. Each LSTM layer contains 512 nodes that use Leaky-ReLU as the activation function. Dropout with a keep probability of $75\%$ is applied to the output of each LSTM layer. The 512-dimensional feature vector obtained from the last LSTM layer is passed to yield predictor.

\begin{figure}
    \centering
    \includegraphics[width=0.4\textwidth]{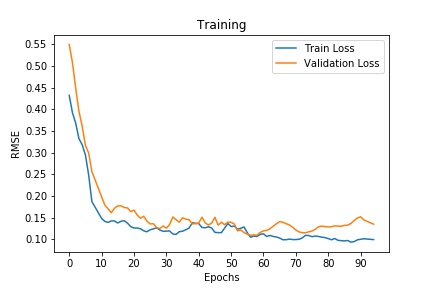}
    \caption{Progression of training and validation error as a function of epochs}
    \label{fig:trainingerror}
\end{figure}

The yield predictor consists of 3 fully connected layers with the first two layers using Leaky-ReLU as the activation function. The yield predictor outputs $\hat{y}_t$ the crop yield in kilograms per hectare for the input sequence until time step $t$. An L2-loss is applied to the predictions corresponding to each time step against the actual output $y_t$. Note that the actual yield at every time step is the same as the yield at the last time step. The overall loss of the entire CNN-LSTM network is defined as follows
\begin{equation}
    Loss = \sum_{t=1}^{24}(\hat{y}_t - y_t)^2
\end{equation}
Applying the L2-loss at each time step increases the flow of gradients to the shared LSTM weights, thus increasing prediction accuracy and faster convergence. This also helps us to predict the yield for intermediate stages of the growing season. Given a test sequence of 24 images, the overall yield is obtained by averaging the yield predicted by the CNN-LSTM model at every time step.
Figure \ref{fig:trainingerror} presents the decrease in the training and validation loss as a function of epochs. We use the model resulting in the lowest validation error for predicting the yield on the test set.

\begin{table*}[t]
\centering
\caption{Comparison of RMSE (in kgs/hectare) for the CNN-LSTM-12 approach against prior and state of the art approaches}
\begin{tabular}{ | p{1.4cm}  | p{1.3cm} | p{1.5cm} | p{1.3cm} |  p{1.3cm} |  p{1.5cm} | p{1.4cm} |  p{1.5cm} | }
\hline
\hline
 State & Decision Forest (NDVI) & Decision Tree (NDVI) & Step Regression (VCI) & Ridge Regression (NDVI) & LSTM + GP (Histogram) & CNN-LSTM-9 & \textbf{CNN-LSTM-12}\\
 \hline
 \hline
 Gujarat & 219 & 259 & 290 & 233 & 140 & 80 & \textbf{48}\\
 Bihar & 835 & 1042 & 775 & 809 & 480 & 460 & \textbf{330}\\
 Haryana & 980 & 1205 & 978 & 1026 & 590 & 234 & \textbf{103.7}\\
 MP & 491 & 602 & 543 & 470 & 370 & 194 & \textbf{161}\\
 UP & 516 & 637 & 509 & 497 & 800 & 138 & \textbf{76}\\
 Rajasthan & 207 & 272 & 222 & 210 & 150 & 117 & \textbf{84}\\
 Punjab & 1065 & 1061 & 1219 & 1061 & 690 & 184 & \textbf{100}\\
 \hline
 \hline 
\end{tabular}
\label{tab:baselinecomparison}
\end{table*}
\section{Experiments and Results}
\subsection{Comparison Against Baselines}
We compare the performance of the proposed model against approaches in the literature that use handcrafted features from the satellite imagery like NDVI and VCI. We train Decision Trees \cite{jonson}, Random Forests, and Ridge Regression models \cite{bolton} using a feature vector of NDVI values derived from each of the 24 satellite images spanning the entire growing season. We also perform step-wise regression with VCI \cite{rice}. We compare our approach against the LSTM+Gaussian Process model \cite{aaai} on the histogram of crop pixels. The parameters for all these approaches were fine-tuned using a cross-validation process. We denote our proposed model that uses raw satellite imagery and contextual information such as water bodies, an agricultural area, and urban landscape as CNN-LSTM-12.

We use root mean square error (RMSE) in kgs/hectare for comparing the performance of the different models. The training set consists of data within the years 2001-2009, the validation set used for tuning the parameters was from the year 2010, and the test set consisted of the data from the year 2011. The results for the different states are presented in Table \ref{tab:baselinecomparison}. It can be observed that the proposed approach performs significantly better than the methods that use NDVI and VCI features by over 70\%. Further, our approach performs better than the LSTM+GP approach of You et al. by over 54\%. We attribute this improvement in the performance of the CNN-LSTM-12 model to its ability to learn features relevant to the task of crop yield prediction, instead of using handcrafted features like a histogram.

\begin{figure*}
    \centering
    \includegraphics[width=0.7\textwidth]{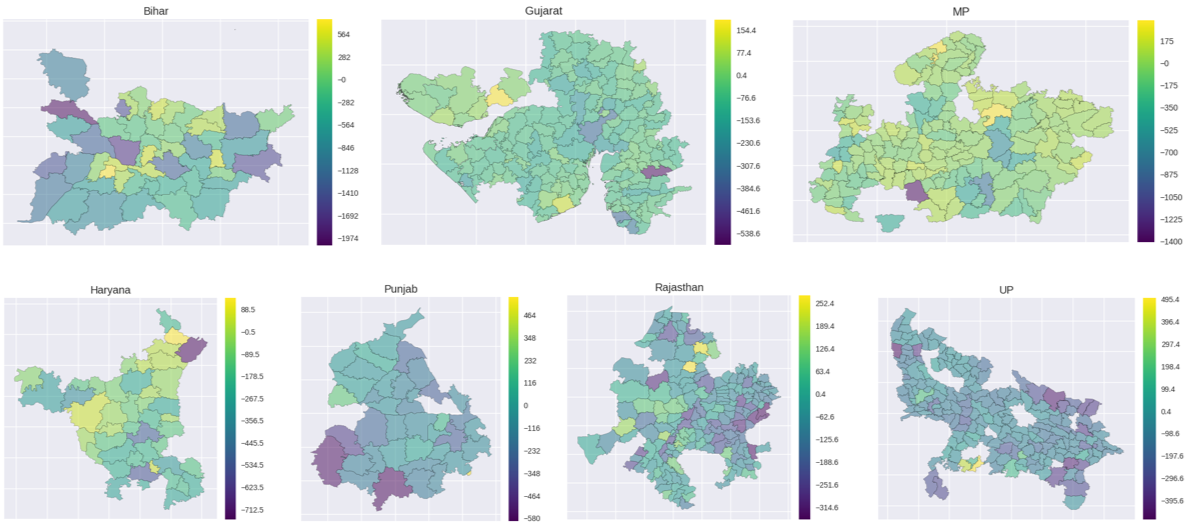}
    \caption{Tehsil level error heat maps for all the 7 states.}
    \label{fig:tehsil-heatmap}
\end{figure*}

\begin{figure*}
    \centering
    \includegraphics[width=0.7\textwidth]{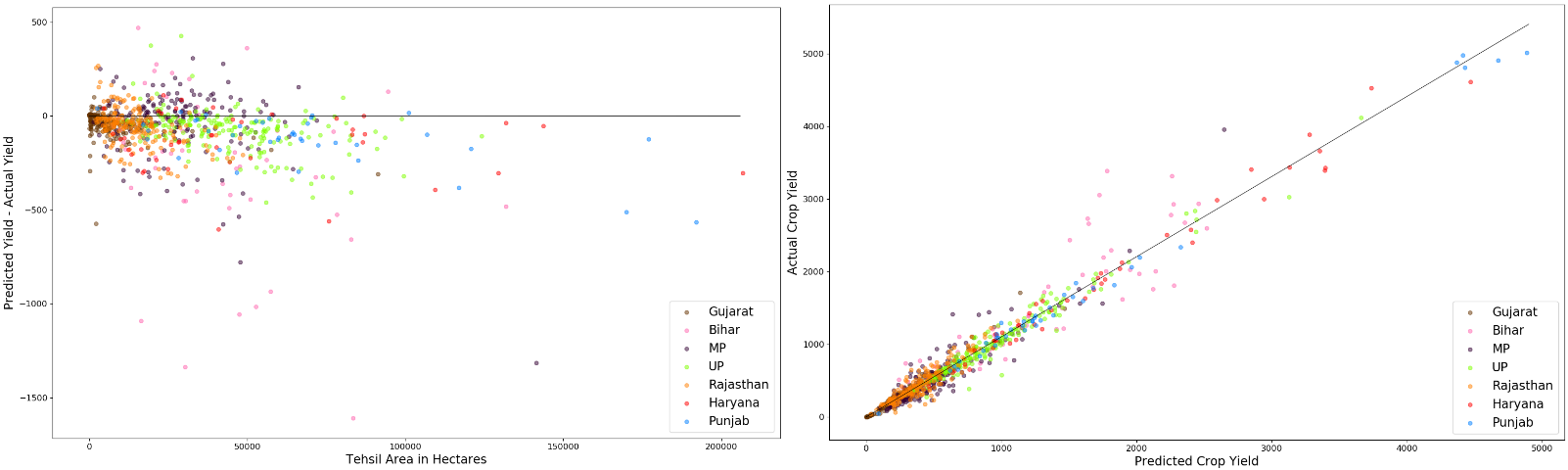}
    \caption{The left figure is the plot of the difference between predicted and actual yield against the area of the tehsils and the plot on the right side illustrates the relationship between predicted and actual yield of all tehsils.}
    \label{fig:tehsil-yield}
\end{figure*}

The crop yield error plots at the tehsil level for every state are presented in Figure \ref{fig:tehsil-heatmap}. It can be observed that for a majority of the tehsils across all the states, the CNN-LSTM-12 model is under-predicting the yield marginally. This is further verified by the plot on the left-hand side in Figure \ref{fig:tehsil-yield}. This compares the error against the size of the tehsil. We observe that for large tehsils, the model always underestimated the yield. However, the number of such large area tehsils is minimal. The relationship between the actual and predicted yield is presented in the right-hand side plot in Figure \ref{fig:tehsil-yield}. This relationship is mostly linear with a slope of $47^o$, indicating that the average performance of the CNN-LSTM-12 model is good.

\begin{figure}
    \centering
    \includegraphics[width=0.5\textwidth]{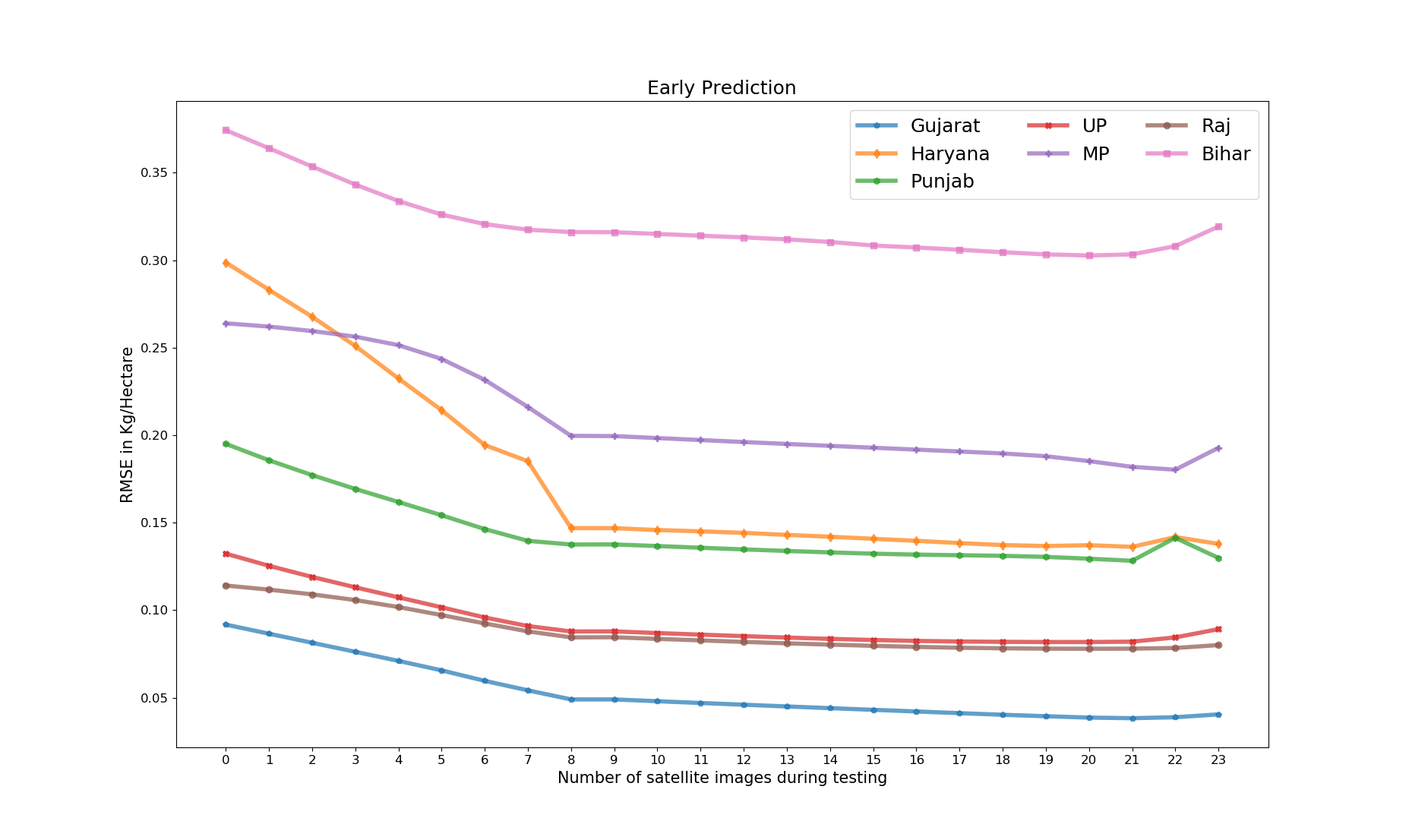}
    \caption{Accuracy of early prediction}
    \label{fig:early-prediction}
\end{figure}
\subsection{Early Crop Yield Prediction}
Another aim of our project is to achieve real-time predictions throughout the growing season. Early crop yield predictions help the government agencies in planning for any contingencies. The CNN-LSTM-12 model has already been trained to predict the yield at every time step. To perform early prediction, we pass only a sub-sequence of the satellite images  $(S_1,S_2....S_t)$ with $t<24$ to the CNN-LSTM-12 model. Figure \ref{fig:early-prediction} shows the performance (RMSE in kgs/hectare) if the prediction was made using only a sub-sequence in an online manner. We observe that the model has a higher error in the early months, as there is not enough information initially on the growth of the plants. However, as more data is made available, we notice an increase in the quality of the prediction for all the states. We notice that the error reduces significantly and consistently at every step until around the $8^{th}$ time step, beyond which there is only a gradual change. This approximately translates to 2 months since the beginning of the sowing season. 

We further observe a slight increase in the error towards the last time step. The final few time steps represent the harvesting part of the crop season. The harvesting is performed over many weeks that is not uniform across and within a tehsil. As a result, we expect to see inconsistencies in the images between areas where the harvesting has been completed and with those where it has not taken place. We suspect this to be the reason for the marginal increase in the error towards the end of the crop season.

\subsection{Importance of Contextual Information}
One of our hypotheses is that integrating contextual information such as the location of water bodies, farmlands, and urban landscape will help the CNN-LSTM-12 model to predict the crop yield more accurately. To test this hypothesis, we train another model without using this information. Specifically, we train a model using only the nine image bands, excluding the last three bands that encode the contextual information. This new model is represented as CNN-LSTM-9. We also mask out regions in these nine bands that do not correspond to agricultural land as encoded in the land use data. The column named CNN-LSTM-9 in Table \ref{tab:baselinecomparison} presents the average RSME for the tehsils of all the states in the study for this model. It is evident that the model that uses information about water bodies, farmlands, and urban landscape performs significantly better (by over 17\%) than the model that does not use this information. This trend is observed across all the states, indicating the importance of the contextual information.
\begin{figure*}
\centering
\includegraphics[width=0.7\textwidth]{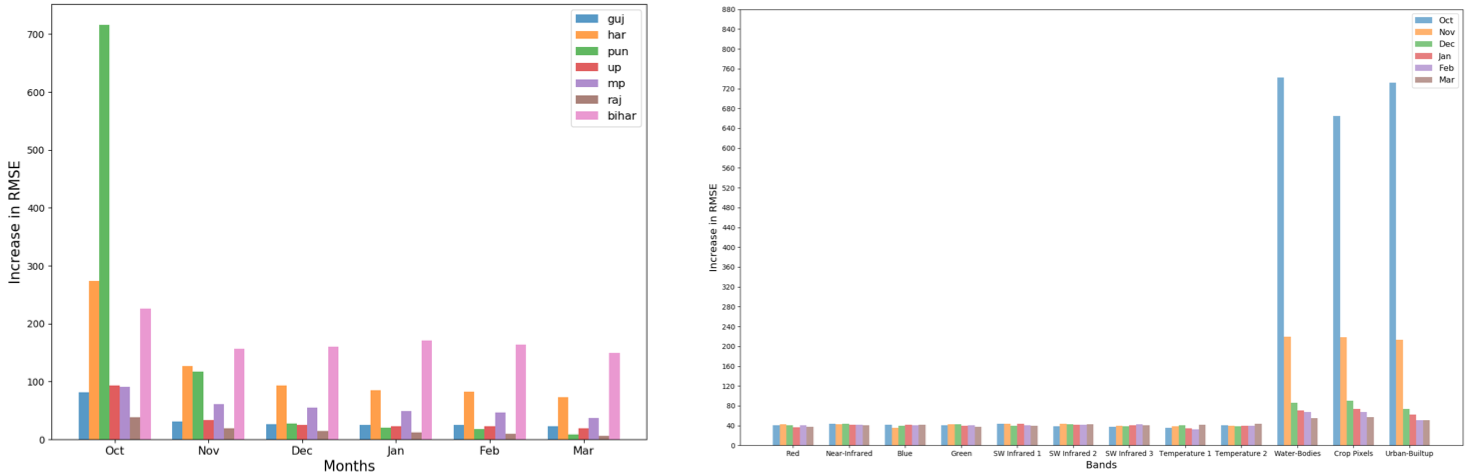}
\caption{Increase in the RMSE when the images of a specific month are replaced with random noise.}
\label{fig:month}
\end{figure*}

\begin{figure}
\includegraphics[width=0.4\textwidth]{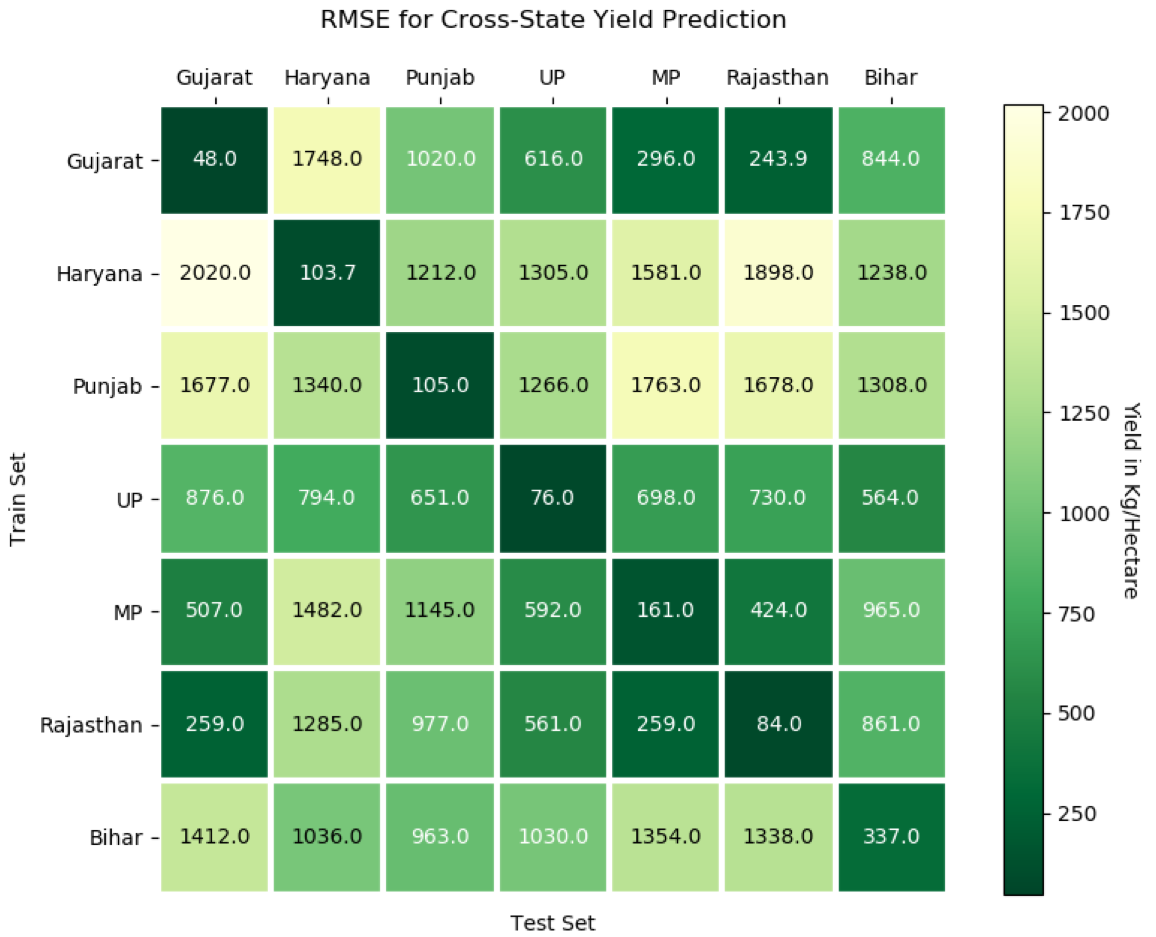}
\centering
\caption{Error of models trained and tested on different states}
\label{fig:gen}
\end{figure}
\subsection{Importance of Image Bands and the Months in the Growing Season}
We perform experiments to analyze how our model is utilizing the input data - the different periods in the growing season and the various bands of the multispectral satellite images for the task of crop yield prediction.

The entire growing season spans six months. Every month, we have four satellite images that are captured approximately every eight days. To analyze the utility of a given month in the growing season, we replace the four images of the month with random Gaussian noise when passing the images to the yield prediction model. We quantify the increase in the RMSE for yield prediction due to this change to estimate the utility of the month in the growing season. The increase in the RMSE for every state and every month is presented in the right-hand side plot in Figure \ref{fig:month}. We observe that the satellite images belonging to the initial month of October are given the maximum importance. This is consistent with the observations in the literature that sowing time is an essential factor in wheat production~\cite{sow-dates}. This further supports our earlier observation on the decrease in the prediction error when the information about the initial two months is made available. As the model sees more satellite images, the increase in the error is only marginal.

We also analyze how the bands are utilized month wise. To see the overall importance given to various bands in the crop yield prediction task, we iteratively send Gaussian noise in place of the individual bands for a given month and observe the increase in the error. 

During October, the model has only seen the first four satellite images of a test data point, which are insufficient for accurate yield prediction. Therefore the model gives maximum importance to bands 10, 11, and 12 signifying the pixels belonging to water bodies, agriculture, and urban built-up. As the model sees more satellite images, it has already recognized the type and the context of each pixel in the sequence of satellite images. Hence, it starts giving lesser importance to the last three bands. The trend that is visible in all the states is that when a band is given significance in a particular month, in the subsequent month, it is immediately given less importance, as the model now gives importance to other bands. The temperature band is consistently given high importance in the later months.

\subsection{Generalizability}
We also try to see how similar different states are in terms of heterogeneity of weather patterns, soil type, farming methods, etc. For this, we train our model on one state and test it on the remaining states. The results are presented in Figure \ref{fig:gen}. We observe that the results when the test state is different are quite poor, with a significant increase in the error (over 1000 in some cases). This further supports our original idea of modeling each state independently.

We had performed a couple of experiments before fixing the state-wise models. A single model was trained using data from 10 bands for all the states; however, the model was having high loss $>$520, while the average loss of state-wise models was $<$120 ($<$100 for some states). The LSTM+GP model on the entire dataset also gave similar losses. 

\section{Conclusion}
We introduce a reliable and inexpensive method to predict crop yields from publicly available satellite imagery. Specifically, we learn a deep neural network model for predicting the wheat crop yield for tehsils in India. The proposed method works directly on raw satellite imagery without the need to extract any hand-crafted features or perform dimensionality reduction on the images. We have created a new dataset consisting of a sequence of satellite images and the exact crop yield for the years 2001-2011 covering a total of 948 tehsils. We use this dataset to train and evaluate the proposed approach on tehsil level wheat predictions. Our model outperforms over existing methods by over 50\%. We also show that incorporating additional contextual information such as the location of farmlands, water bodies, and urban areas helps in improving the yield estimates.

\section{Acknowledgement}
We are grateful to Dr. Reet Kamal Tiwari and  Akshar Tripathi for their inputs and assistance in understanding and collecting the satellite data. We are also grateful to NVIDIA Corporation for supporting this research through an academic hardware grant.


 



%
%
%
{\small
\bibliographystyle{ieee}
\bibliography{wheat-deep-lstm.bib}

\begin{thebibliography}{10}\itemsep=-1pt

\bibitem{lpdaac}
Lp daac.
\newblock \url{https://lpdaac.usgs.gov/}.

\bibitem{unionbudget}
Union budget and economic survey.
\newblock \url{www.indiabudget.gov.in/budget2016-2017/survey.asp}.

\bibitem{FAO}
{“FAO.org.” India at a Glance | FAO in India | Food and Agriculture
  Organization of the United Nations}.

\bibitem{OGD}
{Open Government Data Platform India}.
\newblock \url{https://aps.dac.gov.in/APY/Public_Report1.aspx}, 2018.

\bibitem{albert}
A.~Albert, J.~Kaur, and M.~C. Gonzalez.
\newblock Using convolutional networks and satellite imagery to identify
  patterns in urban environments at a large scale.
\newblock In {\em Proceedings of the 23rd ACM SIGKDD International Conference
  on Knowledge Discovery and Data Mining}, pages 1357--1366. ACM, 2017.

\bibitem{bengio1994learning}
Y.~Bengio, P.~Simard, P.~Frasconi, et~al.
\newblock Learning long-term dependencies with gradient descent is difficult.
\newblock {\em IEEE transactions on neural networks}, 5(2):157--166, 1994.

\bibitem{bolton}
D.~K. Bolton and M.~A. Friedl.
\newblock Forecasting crop yield using remotely sensed vegetation indices and
  crop phenology metrics.
\newblock {\em Agricultural and Forest Meteorology}, 173:74--84, 2013.

\bibitem{sugarcane-fasal}
S.~Dubey, A.~Gavli, S.~Yadav, S.~Sehgal, and S.~Ray.
\newblock Remote sensing-based yield forecasting for sugarcane (saccharum
  officinarum l.) crop in india.
\newblock {\em Journal of the Indian Society of Remote Sensing},
  46(11):1823--1833, 2018.

\bibitem{lstm}
S.~Hochreiter and J.~Schmidhuber.
\newblock Long short-term memory.
\newblock {\em Neural computation}, 9(8):1735--1780, 1997.

\bibitem{jonson}
D.~M. Johnson.
\newblock An assessment of pre-and within-season remotely sensed variables for
  forecasting corn and soybean yields in the united states.
\newblock {\em Remote Sensing of Environment}, 141:116--128, 2014.

\bibitem{kuwata}
K.~Kuwata and R.~Shibasaki.
\newblock Estimating crop yields with deep learning and remotely sensed data.
\newblock In {\em 2015 IEEE International Geoscience and Remote Sensing
  Symposium (IGARSS)}, pages 858--861. IEEE, 2015.

\bibitem{cnn}
Y.~LeCun, L.~Bottou, Y.~Bengio, P.~Haffner, et~al.
\newblock Gradient-based learning applied to document recognition.
\newblock {\em Proceedings of the IEEE}, 86(11):2278--2324, 1998.

\bibitem{rice}
K.~Mallick, J.~Mukherjee, S.~Bal, S.~Bhalla, and S.~Hundal.
\newblock Real time rice yield forecasting over central punjab region using
  crop weather regression model.
\newblock {\em Journal of Agrometeorology}, 9(2):158--166, 2007.

\bibitem{mikolov2010recurrent}
T.~Mikolov, M.~Karafi{\'a}t, L.~Burget, J.~{\v{C}}ernock{\`y}, and
  S.~Khudanpur.
\newblock Recurrent neural network based language model.
\newblock In {\em Eleventh annual conference of the international speech
  communication association}, 2010.

\bibitem{kdd}
B.~Oshri, A.~Hu, P.~Adelson, X.~Chen, P.~Dupas, J.~Weinstein, M.~Burke,
  D.~Lobell, and S.~Ermon.
\newblock Infrastructure quality assessment in africa using satellite imagery
  and deep learning.
\newblock In {\em Proceedings of the 24th ACM SIGKDD International Conference
  on Knowledge Discovery \& Data Mining}, pages 616--625. ACM, 2018.

\bibitem{tushar}
S.~M. Pandey, T.~Agarwal, and N.~C. Krishnan.
\newblock Multi-task deep learning for predicting poverty from satellite
  images.
\newblock In {\em Thirty-Second AAAI Conference on Artificial Intelligence},
  2018.

\bibitem{anoop}
A.~K. Prasad, L.~Chai, R.~P. Singh, and M.~Kafatos.
\newblock Crop yield estimation model for iowa using remote sensing and surface
  parameters.
\newblock {\em International Journal of Applied Earth Observation and
  Geoinformation}, 8(1):26--33, 2006.

\bibitem{ray14a}
S.~S. Ray, S.~Mamatha~Neetu, and S.~Gupta.
\newblock Use of remote sensing in crop forecasting and assessment of impact of
  natural disasters: operational approaches in india.
\newblock {\em Crop monitoring for improved food security}, 2014.

\bibitem{population}
C.~Robinson, F.~Hohman, and B.~Dilkina.
\newblock A deep learning approach for population estimation from satellite
  imagery.
\newblock In {\em Proceedings of the 1st ACM SIGSPATIAL Workshop on Geospatial
  Humanities}, pages 47--54. ACM, 2017.

\bibitem{xie}
M.~Xie, N.~Jean, M.~Burke, D.~Lobell, and S.~Ermon.
\newblock Transfer learning from deep features for remote sensing and poverty
  mapping.
\newblock In {\em Thirtieth AAAI Conference on Artificial Intelligence}, 2016.

\bibitem{aaai}
J.~You, X.~Li, M.~Low, D.~Lobell, and S.~Ermon.
\newblock Deep gaussian process for crop yield prediction based on remote
  sensing data.
\newblock In {\em Thirty-First AAAI Conference on Artificial Intelligence},
  2017.

\bibitem{sow-dates}
M.~Zubair and A.~A. Khakwani.
\newblock Effect of sowing dates on the yield and yield components of different
  wheat varieties.
\newblock {\em Journal of Agronomy}, 5(1):106--110, 2006.

\end{thebibliography}
}

\end{document}